\documentclass[10pt,twocolumn]{article} 
\usepackage{simpleConference}
\usepackage{times}
\usepackage{graphicx}
\usepackage{amssymb}
\usepackage{url,hyperref}
\usepackage{algorithm}
\usepackage{breqn}
\usepackage{algpseudocode}
\usepackage{tabularx}
\usepackage{caption}
\usepackage{float}
\usepackage{mathrsfs,amsmath}
\usepackage[dvipsnames]{xcolor}
\usepackage[activate={true,nocompatibility},final,tracking=true,kerning=true,spacing=true,factor=1100,stretch=10,shrink=10]{microtype}

\begin{document}

\title{Shape Conditioned Human Motion Generation with Diffusion Model}

\author{Kebing Xue, Hyewon Seo\\[\bigskipamount]
ICube laboratory, CNRS--University of Strasbourg, France}
\date{}

\maketitle
\thispagestyle{empty}

\begin{abstract}
Human motion synthesis is an important task in computer graphics and computer vision. 
While focusing on various conditioning signals such as text, action class, or audio to guide the generation process, most existing methods utilize skeleton-based pose representation, requiring additional skinning to produce renderable meshes.
Given that human motion is a complex interplay of bones, joints, and muscles, considering solely the skeleton for generation may neglect their inherent interdependency, which can limit the variability and precision of the generated results. 
To address this issue, we propose a Shape-conditioned Motion Diffusion model (SMD), which enables the generation of motion sequences directly 
in mesh format, conditioned on a specified target mesh. In SMD, the input meshes are transformed into spectral coefficients using graph Laplacian, to efficiently represent meshes. Subsequently, we propose a Spectral-Temporal Autoencoder (STAE) to leverage cross-temporal dependencies within the spectral domain. Extensive experimental evaluations show that SMD not only produces vivid and realistic motions but also achieves competitive performance in text-to-motion and action-to-motion tasks when compared to state-of-the-art methods. 
\end{abstract}

\begin{figure*}[t]
    \centering
    \includegraphics[width=17.5cm]{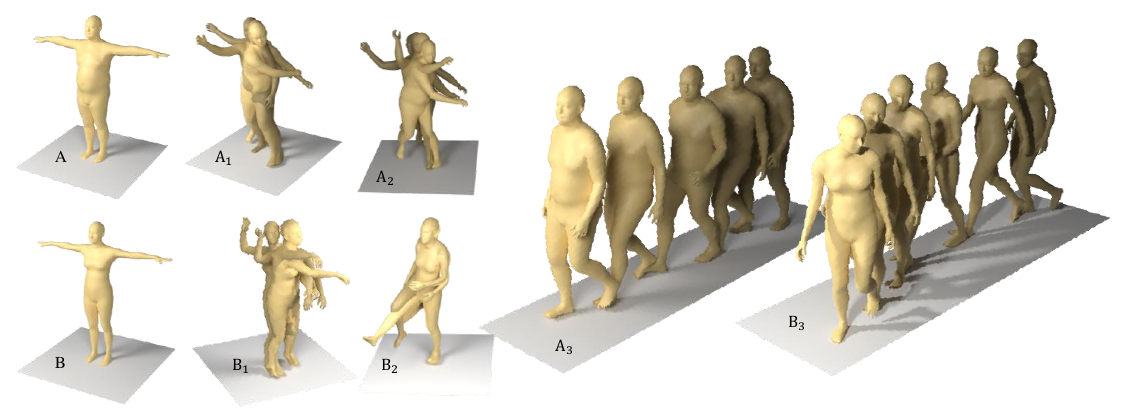}
    \caption{Shape conditioned generation results of SMD. $A_1$, $A_2$, and $A_3$ are based on $A$. $B_1$, $B_2$, and $B_3$ are generated when $B$ is the target mesh.}
    \label{fig:shape_cond}
\end{figure*}

\section{Introduction}
\label{sec:1}
Human motion generation aims to generate realistic and lifelike human movements, which is essential for applications involving virtual characters, such as film production or virtual reality experiences. However, this task poses significant challenges. Firstly, the array of possible motions and body dynamics is extensive and diverse. Secondly, the availability of datasets is limited due to the necessity of professional equipment for motion capture. 
Compounding this challenge is the inherent variability in human motion, even within a single motion class, such as walking or running, the execution can vary substantially from person to person due to subtle anatomical variation and shape differences.
Recently, many studies have been concentrating on this objective, among which the diffusion model\cite{DDPM,improvedDDPM} has been prominently deployed as a backbone. These models can perform motion generation conditioned by various types of signals, such as motion classes, text descriptions\cite{MDM,motiondiffuse}, music\cite{dabral2023mofusion}, and trajectories\cite{guidedMD}. 
All these methods utilize skeleton-based representation as the primary means of depicting body pose and characteristics. This representation condenses human motion into a sequence of joint positions, rotations, and velocities across various body parts \cite{humanML3d},
which provides a simplified and efficient way to represent complex human motion. 
However, it lacks the fine details captured by mesh-based representations, such as muscle deformations.
Since human motion is 
driven by both bones and muscles, generating human motion without considering body shape disregards the natural interdependency between these two elements. This limitation may hinder the model's generalization ability and restrict the variation and precision of the results.  
Moreover, to generate motion for a 3D character, these methods typically require an additional skinning process to attach a renderable skin to an underlying articulated skeleton, often followed by postprocessing steps such as retargeting.
Specifically, the body shape parameters of SMPL model\cite{SMPL} are regressed to the generated skeleton, then combined with the skeleton pose to produce a mesh sequence.
Setting aside the issue of additional computational cost, 
the parameter-based regression process makes it difficult to precisely control the desired body shape. 
These limitations consequently constrain the applicability of these models in 3D animation.

To address the aforementioned challenges, we introduce SMD, a shape-conditioned human motion diffusion framework. SMD is designed to generate realistic human motion directly in renderable mesh format, covering varied motion classes (Figure \ref{fig:shape_cond}). 
Notably, the generated motion is conditioned by a given body mesh, ensuring consistency not only in identity shape throughout the motion but also in the inherent characteristics of the motion itself.
Inspired by SAE\cite{lemeunier2022specRepresentation} for effective information encoding from meshes into deep learning models, we also employ the Laplacian representation to compress human body meshes in the spectral domain. This approach circumvents the great computational and storage costs associated with directly generating the triangle meshes using deep learning models.
To effectively capture temporal dependencies between frames in a motion, we employ a transformer backbone\cite{transformer}. This architecture can also 
aid in integrating information extracted from the conditioning signal into motion features.
Furthermore, SMD enables the bi-modal conditioning of the generation process using both natural language and body shapes. We adopt a classifier-free approach\cite{ho2022classifierfree}, which facilitates a balance between variability and fidelity. This approach eliminates the need for training auxiliary classifiers by sampling both conditional and unconditional instances from the same model.
Additionally, to make the shape conditioning process more user-friendly, SMD accepts meshes in any pose as the target mesh, with the help of a shape encoder pre-trained with a contrastive loss\cite{he2020momentum}, which is capable of extracting shape identity features from meshes in arbitrary poses. Integrating it into our framework allows for precise control of the generated body shape.  

Finally, to validate our model, we perform extensive qualitative and quantitative experiments on BABEL\cite{BABEL:CVPR:2021} and HumanML3d\cite{humanML3d}, both of which are the annotations of AMASS\cite{AMASS} dataset. The results show that our proposed SMD achieves competitive performance compared with state-of-the-art works on several tasks (text-to-motion and action-to-motion). We also show that our approach reliably generates high-quality and vivid human motion while maintaining the body shape consistent with a given target mesh.
 
We summarize our contributions as follows: 

\noindent 1. We introduce a novel framework capable of directly generating human motion in the form of mesh sequences;

\noindent 2. We propose to leverage human body mesh as a conditioning signal, and generate realistic human motion based on it; 

\noindent 3. We evaluate our approach extensively across multiple tasks, demonstrating competitive performance in each scenario.

\section{Related Work}

\label{sec:2}

\textbf{Human motion synthesis} has become a long-standing research topic\cite{bowden2000learning}. By applying deep learning methods to this field, researchers first perform unconstrained generation, whose target is to generate vivid and realistic human motion sequences without any constraints\cite{fragkiadaki2015recurrent,li2017auto,pavllo2018quaternet}. With the advancement of deep learning methods and increasing demands, some works tried to guide motion generation with different action classes. Action2Motion\cite{guo2020action2motion} uses conditional temporal VAE to capture frame-level features. ACTOR\cite{petrovich2021action} adopts conditional transformer VAE to better analyze temporal dependency between frames. ActFormer\cite{xu2023actformer} uses a GAN-based transformer to realize multi-person motion generation. With the development of pre-trained large language models such as CLIP\cite{radford2021clip} and RoBERTa\cite{liu2019roberta}, they provided a bridge between human intention and machine understanding and showed new possibilities to control generative deep learning models. Text-conditioned motion generation thus begins to dominate research frontiers\cite{MDM,kim2023flame,motiondiffuse,dabral2023mofusion}.

\textbf{Diffusion models} have achieved significant success in various tasks such as image denoising, inpainting, super-resolution, and image generation. The work of Sohl-Dickstein \textit{et al}.\cite{sohl2015deep}
first introduce this theory from non-equilibrium statistical physics into the deep learning field. Based on it, Ho \textit{et al}. \cite{ho2020ddpm} design DDPM for high-quality image synthesis,  Others try to augment the efficacy and fidelity of diffusion model\cite{song2020ddim,improvedDDPM}. 
Considering the strong generalize ability of the diffusion model, Dhariwal and Nichol\cite{diffbeatgan} further apply it in class-conditioned image generation tasks, by designing classifier guidance, the diffusion model first surpasses GAN\cite{goodfellow2020gan} on this task. To avoid training auxiliary classifiers, Ho and Salimans proposed a classifier-free guidance\cite{ho2022classifierfree}. The significant success in applying the diffusion model in the field of image generation demonstrates its remarkable generalization ability and encourages researchers to apply it in other tasks such as audio synthesis\cite{popov2021grad,kong2020diffwave} and point cloud generation\cite{luo2021diffusion}.

\textbf{Diffusion-based motion generation} has thus drawn considerable interest from researchers.
MDM\cite{MDM}, Flame\cite{kim2023flame}, MotionDiffuse\cite{motiondiffuse} first try to apply the diffusion model in human motion generation tasks, they use pre-trained large language models to encode commands described in natural language, then guide the generation process based on these feature of text to generate motions follow description. Some of them have also tried motion inpainting. Mofusion\cite{dabral2023mofusion} realizes a music-conditioned generation. 
Make-An-Animation\cite{make-an-animation} extracts a large-scale pseudo-pose dataset from image-text datasets, enhancing the stability of the generation process through pre-training the model with this dataset.
MLD\cite{chen2023executing}, MultiAct\cite{lee2023multiact}, EMS\cite{qian2023breaking} and Fg-T2M\cite{wang2023fg-t2m} focus on generating a longer motion sequence with finer-grained text descriptions.
AttT2M\cite{zhong2023attt2m} propose to use a multi-perspective attention mechanism to extract dependency between body parts and different frames to augment the generation quality.
Interdiff\cite{xu2023interdiff} uses the diffusion model to simulate human-object interactions.
GMD\cite{guidedMD} realize a motion generation along given trajectories or avoid certain obstacles.
PhysDiff\cite{yuan2023physdiff} proposes to apply motion correction based on a physic simulator iteratively during denoising steps of reverse diffusion to generate physically plausible motions.

\begin{figure*}[t]
\includegraphics[width=17.5cm]{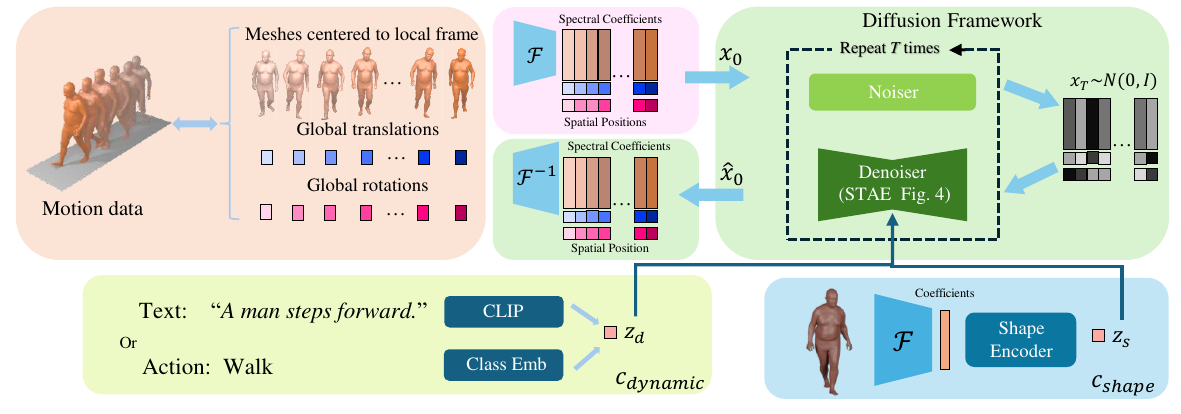}
\caption{Method overview: 1) From each mesh in the motion data, the vertex coordinates in the local frame are transformed into spectral coefficients by using a graph Fourier transformation; 2) The coefficients together with the rotations and translations are used to train the diffusion model; 3) After training, Spectral-Temporal Autoencoder (STAE) generates motion from a random noise, conditioned on a conditioning signal $z_d$ and a target mesh embedding $z_s$, by denoising it iteratively.}
\label{fig:Overview}
\centering
\end{figure*}

\section{Method}
\label{sec:3}
We introduce SMD, a diffusion-based framework for shape-conditioned human motion generation.
\subsection{Method Overview}
\label{sec:3.1}
An overview of our framework is shown in Figure \ref{fig:Overview}, given a conditioning target body mesh $c_{shape}$ in an arbitrary pose and a conditioning signal $c_{dynamic}$ in the form of text or action class, our objective is to generate a motion sequence $M^{1:F}$ of length $F$ that corresponds to the description and under the same body shape as the target mesh. In our work, we use body meshes that are compatible with SMPL\cite{SMPL}. For training, starting from a motion sequence $M^{1:F} \in \mathbb{R}^{F \times N \times 3}$ consisting of $F$ triangle meshes each with $N$ vertices, we first separate the global position $P^{1:F}\in \mathbb{R}^{F\times 3}$ and rotation $R^{1:F}\in \mathbb{R}^{F\times 3}$ of the root joint for each frame. This will ensure that the meshes are centered at the origin point and facing in the same direction. Next, the centered meshes are transformed into spectral coefficients $C^{1:F} \in \mathbb{R}^{F \times k \times 3}$ by applying the graph Fourier transform, where $k$ is the number of coefficients. These coefficients are further normalized by subtracting the mean and dividing by the standard deviation, computed across the entirety of the training dataset.
The separated translation and rotation of the root joint are concatenated with the normalized spectral coefficients to form $x^{1:F}_0 \in \mathbb{R}^{F \times (k+2) \times 3}$, which is used to train the diffusion model, along with features extracted from the condition signals.

The conditioning signals consist of two kinds: one is utilized to constrain the body shape, while the other guides the dynamics of the generated motion. Specifically, given a target mesh $c_{shape}$ and an action class or textual description $c_{dynamic}$, our SMD generates 
a deforming body mesh exhibiting the desired motion and that maintains the same body identity as the target. While our current model is designed to support meshes in SMPL topology, any arbitrary topology can be made SMPL-compatible through the use of surface correspondence or registration techniques.

During inference, we sample $x^{1:F}_T$ from a normal distribution $N(0,I)$ and iteratively denoise it into $\hat{x}^{1:F}_0$ guided by $c_{dynamic}$ and $c_{shape}$ with the trained STAE. Then a reverse graph Fourier transform is applied to the generated coefficients to cover them back into meshes. Finally, spatial position and rotation are injected into them to build a motion in 3D.

\subsection{Mesh Spectral Representation}
\label{sec:3.2}
We use graph Laplacian to transform meshes into spectral coefficients and use them as the model's input for training\cite{lemeunier2022specRepresentation}. A human body triangle mesh $M$ can be considered as a graph with $N$ nodes. The position of vertices can be expressed as $f(i)=(x_i,y_i,z_i),i\in [1, N]$ where $x,y,z$ correspond to the 3D coordinates of vertices. The adjacency matrix $A$ of this graph is a square matrix whose elements indicate whether pairs of vertices are adjacent or not and its degree matrix $D$ is a diagonal matrix whose elements correspond to the number of edges attached to each vertex. The Graph Laplacian $L$ is defined as $L=D-A$. Since all meshes we use share the same topology introduced in SMPL\cite{SMPL}, their graph Laplacians are also the same.

Let $\{\lambda_l\}_{l=0,1,...,F-1}$ and $\{\textbf{u}_l\}_{l=0,1,...,F-1}$ be the eigenvalues and eigenvectors of $L$ satisfying $L\textbf{u}_l=\lambda_l \textbf{u}_l$, similar to the classical Fourier transform, we can represent $M$ in spectral coefficients by applying the graph Fourier transform $\mathscr{F}$:
\begin{equation}
    \mathscr{F}(\lambda_l)=\left< \textbf{f},\textbf{u}_l \right> =\sum_{i=1}^Nf(i)u_l^*(i),
\end{equation}
and the reverse graph Fourier transform can be expressed as:
\begin{equation}
    f(i) =\sum_{l=0}^{N-1}\mathscr{F}(\lambda_l)u_l(i).
\end{equation}
Since the graph Laplacian $L$ is a real symmetric matrix, its eigenvalues are all non-negative, we sort them in a decreasing order. Eigenvectors associated with lower eigenvalues vary slowly across the graph, while eigenvectors associated with higher eigenvalues are more likely to have different values on adjacent nodes. And the most important part of information about the human body is contained in the low frequencies.

\begin{figure}
    \includegraphics[height=5cm]{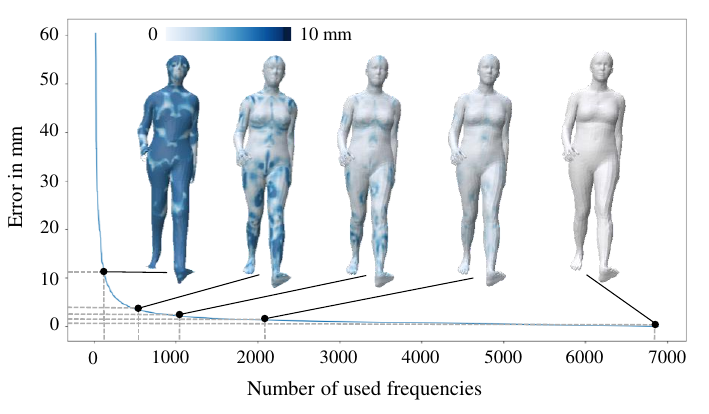}
    \caption{Reconstruction error as a function of the number of used eigenvectors. SMPL meshes are compressed by applying graph Fourier transform/inverse transform using a certain number of eigenvectors, from left to right we try 128,512,1024,2048 and 6890 eigenvectors, the color corresponds to the distance between the compressed mesh and original mesh.}
    \label{fig:mesh spectral representation}
\end{figure}

We choose $k$ eigenvectors where $k\ll n$ corresponds to $k$ lowest eigenvalues to calculate the coefficients, this can reduce the size of input training data, and further reduce the computational complexity of our method. Of course, this process will lose some information, but as shown in Figure \ref{fig:mesh spectral representation}, by choosing a proper $k$, we achieve a high level of representation accuracy while minimizing information loss to a great extent.

Compared to using graph convolutional networks which also exploit the information of graph in spectral domain by convolving along the graph topology, this approach avoids downsampling/upsampling, which can potentially hinder the exploitation of spectral information to its fullest extent\cite{lemeunier2022specRepresentation}.

\begin{algorithm}
\caption{Sampling}
\label{sampling}
  \begin{algorithmic}[1]
  \State \textbf{Input:} conditions $C$, well trained $STAE$
  \State $x_t\sim N(\textbf{0},\textbf{I})$
  \For{t=T,...,1}
    \State $\textbf{z}\sim  N(\textbf{0},\textbf{I})$ if $t>1$, else $\textbf{z}=0$
    \State $x_{t-1}=\tilde{\mu}_t(x_t,STAE(x_t,t,C))+\delta_t\textbf{z}$
  \EndFor
  \State $\textbf{return} x_0$

  \end{algorithmic}
\end{algorithm}

\subsection{Motion diffusion model}
\label{sec:3.3}
The overview of our diffusion model is shown in Figure \ref{fig:Overview}. It is based on the diffusion framework, in which we design a Spatial-Temporal Autoencoder as the denoiser for the reverse process.\newline

\textbf{Diffusion model} is a deep generative model that consists of two stages, a forward diffusion stage which acts as a Markov noising process, and a backward denoising process based on a deep learning model. In the forward process, given an uncorrupted training sample $x^{1:F}$, the noise version $x^{1:F}_1$, $x^{1:F}_2$...,$x^{1:F}_t$ at diffusion step $t$ can be driving from the following approximate posterior:
\begin{equation}
    q(x^{1:F}_T)=\prod_{t=1}^Tq(x^{1:F}_t|x^{1:F}_{t-1}),\forall t \in {1,...,T},
    \label{diffusion poste}
\end{equation}
\begin{equation}
    q(x^{1:F}_t|x^{1:F}_{t-1})=N(x^{1:F}_t;\sqrt{1-\beta_t}x^{1:F}_{t-1}, \beta_t \textbf{I}), 
    \label{diffusion1}
\end{equation}
where $T$ corresponds to the predefined maximum diffusion step and $\beta_1,...\beta_t \in [0,1)$ are the predefined noise schedule. For simplicity, we use $x_t$ to denote the full sequence $x_t^{1:F}$
at diffusion step $t$ and use $x_0$ for the uncorrupted input. An important property of this process is that a sample at any step $t$ is driven directly from the input $x^{1:N}$ with this formula\cite{DDPM}:
\begin{equation}
    q(x_t|x_0)=N(x_t;\sqrt{\hat{\alpha_t}}x_0, (1- \hat{\alpha_t})\textbf{I}),
    \label{direct diffusion}
\end{equation}
where $\hat{\alpha_t}=\prod^t_{i=1}\alpha_t$ and $\alpha_t=1-\beta_t$. This allows us to easily train a denoising model for arbitrary steps by defining a noise schedule $\{ \beta_t \}_{t=1,...,t}$. Following the setting in this paper \cite{improvedDDPM}, we adopt the cosine noise schedule in terms of $\hat{\alpha_t}$ as:
\begin{equation}
    \hat{\alpha_t}=\frac{f(t)}{f(0)}, f(t)=cos(\frac{t/T+s}{1+s} \cdot \frac{\pi}{2})^2.
\end{equation}
When $t$ is closer to $T$, $\hat{\beta_t}$ will be close to 1, in which case we can approximate $p(x_T) \sim N(0, I)$.

The goal of the conditioned motion diffusion model is to generate a sequence of human motion under certain constraints, which is further abstracted in this formula $p(x_0|C)$ in which $C$ corresponds to condition signals. This is realized by using a deep learning model whose parameter is represented as $\theta$ to denoise an initial state $x_T$ sampled from a normal distribution into an uncorrupted sample by sampling from these distributions iteratively:
\begin{equation}
    p_\theta(x_0|C)=p(x_T)\prod_{t=1}^T p_\theta(x_{t-1}|x_t,C).
    \label{eqn:reverse}
\end{equation}

Some works applied the diffusion model to the image generation task\cite{DDPM,diffbeatgan}, their model is trained to predict the noise during training. In our context, predicting the input $x_0$ during training will allow us to apply geometric losses more easily, making the generated motion more stable. Our training objective for the diffusion model can be written as the following expected value:
\begin{equation}
    L_{diff}=E_{x_0\sim q(x_0|C),t\sim[1,T]}[{||x_0-STAE(x_t,t,C)||}^2].
\end{equation}

The probability distribution for the denoise process in Equation \ref{eqn:reverse} can be simulated by calculating its mean and variance by the deep learning model:
\begin{equation}
        p_\theta(x_{t-1}|x_t,C)=N(x_t-1;\mu_\theta(x_t,t,C),\Sigma_\theta(x_t,t,C))
        \label{reverse}
\end{equation}
Following the setting in DDPM \cite{DDPM}, we fix the variance term $\Sigma_\theta(x_t,t,C)$ to $\delta_t^2\textbf{I}$ where $\delta_t^2=\hat{\beta}_t=\frac{1-\hat{\alpha}_{t-1}}{1-\hat{\alpha}_t}$ for a more stable training. 
To transform the predicted $\hat{x}_0$ to $\mu_\theta(x_t,t,C)$, we need to consider the forward process posteriors which are tractable given $x_0$ \cite{DDPM}:

\begin{equation}
    q(x_{t-1}|x_t,x_0)=N(x_{t-1};\tilde{\mu_t}(x_t,x_0), \tilde{\beta_t}\textbf{I}), 
\end{equation}

\begin{equation}
    \tilde{\mu}_t(x_t,x_0)=\frac{\sqrt{\hat{\alpha}_{t-1}}\beta_t}{1-\hat{\alpha}_t}x_0+\frac{\sqrt{\alpha_t}(1-\hat{\alpha}_{t-1})}{1-\hat{\alpha_t}}x_t.
\end{equation}
We thus replace the $x_0$ with the predicted $\hat{x}_0=STAE(x_t,t,C)$ in $\tilde{\mu}_t(x_t,x_0)$ to approximate the $\mu_\theta$ in Eq. \ref{reverse} . Sampling $x_{t-1}\sim p_{\theta}(x_{t-1}|x_t,C)$ is thus performed via a parametrization trick by calculating $x_{t-1}=\tilde{\mu}_t(x_t,\hat{x}_0)+\delta_t\textbf{z}$, where $\textbf{z}\sim N(\textbf{0},\textbf{I})$, as shown in Algorithm \ref{sampling}.\newline

\begin{figure}
    \includegraphics[height=4.6cm]{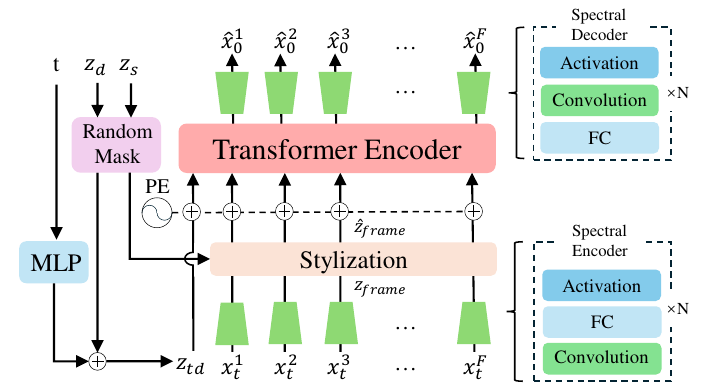}
    \caption{Overview of our Spectral-Temporal Autoencoder (STAE).}
    \label{fig:STAE}
\end{figure}

\textbf{Spectral-Temporal Autoencoder} consists of a spectral encoder, a spectral decoder, and a Transformer encoder, as illustrated in Figure \ref{fig:STAE}. 
Given that the conditioning signals come from different modalities, with the dynamic conditions being either an action class or a text description, and the shape condition being represented by a triangle mesh, we adopt distinct strategies to integrate them. Note that this is different from other approaches that sum different conditioning signals into one before sending it to the model. The embedding of diffusion step $t$ and features extracted from dynamic conditions $z_d$ are summed together to $z_{td}$. It is concatenated to the first place of the features extracted from spectral coefficients with the spectral encoder. The subsequent transformer encoder allows it to have different influences on different temporal parts of motion. 
Different from $z_d$, the feature $z_s$ extracted from target meshes is expected to have a consistent influence on all frames, so we adopt a stylization block\cite{motiondiffuse} to inject them into motion features. The stylization block consists of three dense layers $\psi_b$, $\psi_w$, and $\phi$, they map the original feature $z_{frame}$ to $\hat{z}_{frame}$ with $\hat{z}=z \circ \psi_w(\phi(z_s))+\psi_b(\phi(z_s))$. 

A transformer encoder is also used to further exploit the temporal dependency between the features of different frames, its self-attention mechanism permits the calculation of features in each frame based on all other frames, which implicitly ensures the continuity and consistency of the whole motion. The output will be decoded using a spectral decoder to match the dimensions of the input. The spectral encoder and decoder are composed of a sequence of convolutional blocks. Each block comprises a convolutional layer, a fully connected layer, and an activation layer, as shown in Figure \ref{fig:STAE}. 

\subsection{Conditioned generation}
\label{sec:3.4}

\textbf{Dynamic conditions} can be an action class or a text description. We adopt CLIP\cite{radford2021clip} as a text encoder to embed the text prompt and use a trainable tensor as embeddings for different action classes. \newline

\textbf{Target mesh} is embedded by a shape encoder capable of embedding the input mesh into features that solely include the identity shape of the body mesh independently of poses. This encoder is trained within a \textit{shape autoencoder} using the paradigm of contrastive learning. It shares the same architecture as the STAE, except that the Transformer Encoder is omitted. Given meshes of different characters in arbitrary poses, it is trained to predict the mesh in canonical T-pose. The contrastive learning loss is adopted to better build a uniform representation space. T-pose meshes $M_i^{t\_in}$ and meshes in arbitrary poses $M_i^{a\_in}$ where $i\in[1,N]$ corresponds to $N$ different chosen characters are sent to the model. The loss function is:
\begin{equation}
    L_{ShapeEmb} = L_{mesh}+L_{contrast},
\end{equation}
\begin{dmath}
    L_{mesh}=\frac{1}{2N}(\sum_{i=1}^N{||M_i^{t\_out}-M_i^{t\_in}||}^2+ \sum_{i=1}^N{||M_i^{a\_out}-M_i^{t\_in}||}^2),
\end{dmath}
\begin{dmath}
    L_{contrast}=-\sum_{i=1}^N\frac{1}{N}log\frac{exp(sim(z_i^{t},z_i^{a})/\tau)}{\sum_{k=1,k\neq i}^Mexp(sim(z_i^{t},z_k)/\tau)},
\end{dmath}
where $M_i^{t\_out}$,$M_i^{a\_out}$ are predicted meshes and $z_i^{t}$,$z_i^{a}$ are latent features. Minimizing $L_{contrast}$ requires maximizing the similarity between $z_i^{t}$ and $z_i^{a}$ which means attracting the features associated with the same character close to each other. At the same time, it pushes the features extracted from different characters away by minimizing the similarity between $z_i^{t}$ and $z_k$.
The $L_{mesh}$ encourages the model to focus on shape information when facing meshes in different poses. These two loss terms will ensure we build a pose-independent representation space for identity shapes. Note that this shape encoder is also used to embed the target mesh $c_{shape}$ into $z_s$. \newline

\textbf{Classifier free guidance} is an approach to achieve highly precise conditioned sampling for a diffusion model without the need for training auxiliary models. Specifically, as shown in Figure \ref{fig:STAE}, we randomly mask a certain proportion of the condition signals to $\emptyset$ to simulate the unconditioned generation. Then, during inference, we can manipulate a ratio $s$ between the conditionally generated result and the unconditionally generated result to trade off diversity and fidelity, formulate as:
\begin{dmath}
    STAE_s(x_t,t,C)=STAE(x,t,\emptyset) +\\s_d \cdot (STAE(x_t,t,c_{dynamic},\emptyset_{shape})-STAE(x_t,t,\emptyset)) +\\s_s \cdot (STAE(x_t,t,\emptyset_{dynamic}, c_{shape})-STAE(x_t,t,\emptyset)).
\end{dmath}

\subsection{Losses}
In our work, for each denoise step, we predict the uncorrupted signal $\hat{x}_0$, which is composed of rotations $R^{1:F}$, translations $P^{1:N}$, and coefficients $C^{1:F}$. As discussed in Section \ref{sec:3.2}, coefficients associated with lower frequencies inherently encapsulate richer information, often characterized by larger values. To amplify our model's focus on these coefficients we multiply the coefficient loss by their respective variances. The coefficient loss is expressed as:
\begin{equation}
    L_{coef}=\frac{1}{F} Var(\tilde{C})\sum^F_{i=1}{||C_0^i-\hat{C}_0^i||}^2,
\end{equation}
where $Var(\tilde{C}) \in \mathbb{R}^{k\times3}$ is the variance of coefficients calculated on the whole training set. Since the coefficients can be further reversed into 3D mesh with pre-calculated eigenvectors, our losses will also be applied to spatial domains:
\begin{equation}
    L_{mesh}=\frac{1}{F}\sum^F_{i=1}{||M_0^i-\hat{M}_0^i||}^2;
\end{equation}
\begin{equation}
    L_{pos}=\frac{1}{F}(\sum^F_{i=1}{||{R}_0^i-\hat{R}_0^i||}^2+\sum^F_{i=1}{||{P}_0^i-\hat{P}_0^i||}^2).
\end{equation}
These three terms keep the coefficients, mesh, and position consistent with the input. To ensure the smoothness of the moving trajectory, we further apply a residual loss to control the translation:
\begin{equation}
    L_{res\_P}=\frac{1}{F-1}\sum^F_{i=2}{||(P_0^i-P_0^{i-1})-(\hat{P}_0^i-\hat{P}_0^{i-1})||}^2.
\end{equation}
Overall, our training loss is:
\begin{equation}
    L=\lambda_d \cdot L_{diff}+\lambda_c \cdot L_{coef}+\lambda_m \cdot L_{mesh}+\lambda_{p} \cdot L_{pos}+\lambda_{r}\cdot L_{res\_P}.
\end{equation}

\section{Experiments}
\label{sec:5}
To evaluate the effectiveness of our method, we utilize SMD in two typical settings: text-to-motion and action-to-motion generation, each combined with a conditioning target mesh.
Furthermore, since our model supports shape-conditioned generation, we evaluate the shape consistency between generated motions and the given target mesh.
\subsection{Settings}
\label{sec:5.1}
\noindent\textbf{Datasets.} We train our model by using HumanML3D\cite{humanML3d} dataset, which contains 32,357 per-motion text annotations on 10,524 motion sequences from the AMASS dataset\cite{AMASS}. AMASS is chosen because it provides mesh data for each motion, which is required by our model. 
We also augment the data by mirroring all sequences about the sagittal plane, following the settings in HumanML3D \cite{humanML3d}. Their test set for the text-to-motion generation is used for the evaluation. For the action-to-motion generation, we use the BABEL\cite{BABEL:CVPR:2021} dataset, which provides per-motion action labels for the AMASS\cite{AMASS} dataset. We chose 8 action classes, resulting in 10,688 motions.

\noindent \textbf{Evaluation metrics.} Following the common settings for text-to-motion generation, we evaluate our SMD on text-to-motion tasks using Fr\'echet Inception Distance(FID), R-Precision, and diversity. FID quantifies the realism and diversity of generated motions by comparing their distribution in latent space with the ground truth distributions using a pre-trained motion encoder. R-precision measures the relevance of the generated motions to the input prompts, and diversity quantifies the variability of the generated motions. 
While our model generates motions in the form of mesh sequences, for comparisons with other works that are mostly skeleton-based, we convert our model's generated mesh sequences into skeletons. This is achieved by using a regression matrix introduced in SMPL\cite{SMPL} to regress joints from the mesh. The aforementioned metrics can then be computed on the regressed skeleton-based motions. 
Furthermore, SMD is trained with mesh geometry loss, which implicitly empowers it to better account for the mesh-environment interaction, and generate physically plausible motions. To evaluate this ability, we adopt three additional 
metrics: \textit{Penetration} measures ground penetration by the body, \textit{floating} measures the extent of mesh floating above the floor, and \textit{skating} measures foot sliding effect, all in millimeters (\textit{mm}). For the action-to-motion generation, \textit{accuracy} replaces the R-Precision by calculating the classification accuracy of a pre-trained classifier on the generated motion. To measure the performance of shape-conditioned generation, we measure the \textit{consistency} of identity shape between the generated motion and the conditioning target mesh by computing the positional error between corresponding vertices of their respective pose-normalized meshes. 

\noindent \textbf{Implementation details.} The main hyperparameter values of the SMD are shown in Table \ref{hyperparameters}. It took less than three days of training on a single NVIDIA GeForce RTX 3080 GPU. 
\begin{table}
\centering
\caption{Hyperparameters.} 
\label{hyperparameters} 
\begin{tabular}{c c}
\hline
Configuration & Value \\
\hline
$k$ & 1024\\
Optimizer & Adam \\
Learning rate & 1e-4\\
Number of multi-head attention & 8\\
Latent dimension & 256\\
Dropout & 0.2 \\
$\lambda_d, \lambda_c, \lambda_m$ & 1, 1, 1 \\
$\lambda_p$ & 50\\
$\lambda_r$ & 1e4\\
\textit{k}: Number of frequencies used &1024\\
Batchsize & 32 \\
$s_d$ & 0.85 \\
$s_s$ & 0.7 \\
\hline
\end{tabular}
\end{table}

\begin{table}[!htbp]
\caption{Action-to-motion results on BABEL\cite{BABEL:CVPR:2021} dataset. $\uparrow$ means the larger values are better, and $\downarrow$ indicates the smaller values are better.}
\label{table:actmotion}
\centering
\begin{tabular}{ccc}
\hline
 Method&FID$\downarrow$&Accuracy$\uparrow$\\
 \hline
 Real(Skeleton)&0.001&0.984\\
 Real(Mesh)&0.001&0.997\\
 MDM\cite{MDM}&0.173&0.979\\
 SMD(Skeleton)&0.251&0.974\\
 SMD(Mesh)&\textbf{0.161}&\textbf{0.991}\\
 \hline
\end{tabular}
\end{table}

\begin{table*}[!h]
\caption{Text-to-motion results on HumanML3D\cite{humanML3d} dataset. $\rightarrow$ means the results are better if the value is closer to the real distribution.}
\label{textmotion}
\centering
\begin{tabular}{ccccccc}
\hline
 Method&FID$\downarrow$&R-Precision$\uparrow$&Diversity$\rightarrow$&Penetrate$\downarrow$&Float$\downarrow$&Skate$\downarrow$\\
 \hline
 Real&0.002&0.797&9.50&5.965&2.354&0.929\\
 T2M\cite{humanML3d}&1.067&0.740&9.188&11.897&7.779&2.908\\
 MDM\cite{MDM}&0.489&0.707&9.45&11.291&18.876&\textbf{1.406}\\
 MotionDiffuse\cite{motiondiffuse}&0.630&0.782&9.410&20.278&6.450&3.925\\
 Fg-T2M\cite{wang2023fg-t2m}&0.243&\textbf{0.783}&9.278&-&-&-\\
 \hline
 SMD(Ours)&\textbf{0.214}&0.737&\textbf{9.472}&\textbf{8.741}&\textbf{5.854}&2.576\\
 \hline
\end{tabular}
\end{table*}

\subsection{Shape conditioned generation}
\begin{figure}
    \centering
    \includegraphics[width=8.5cm]{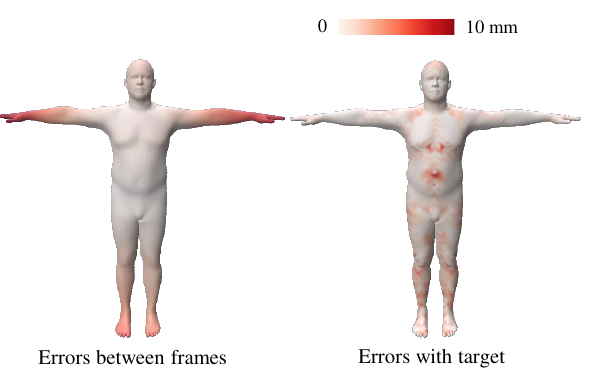}
    \caption{Average errors in shape consistency within the motion (left) and in comparison with the target (right).}
    \label{shape error}
\end{figure}

Our model can generate motion conditioned on a target mesh. We assess the performance of this task by measuring the shape consistency within different frames of the same generated motion and between the generated and target meshes. As introduced in Section \ref{sec:3.4}, our shape embedder is trained to transform a mesh in an arbitrary pose into the canonical T-pose. Utilizing this model, we convert the meshes in the generated motion into T-poses and calculate the shape error based on the Euclidean distance between corresponding vertices in these meshes. We randomly select 20 characters, with each character represented by five meshes in various poses, and use these to generate 1,000 random sequences. The average inner-sequence identity shape error measures at $1.19\pm0.68$ \textit{mm}, while the discrepancy between the target mesh and the generated motion stands at $2.34\pm1.3$ \textit{mm}. In comparison, the compression error arising from the limited number of eigenvectors used for the graph Fourier transform is $0.75\pm0.58$ \textit{mm}. The error distribution is shown in Figure \ref{shape error}: The inter-sequence identity shape errors concentrate on foot and hand while the errors with respect to the target follow a similar distribution to the reconstruction error shown in Figure \ref{fig:mesh spectral representation}. These quantitative results show our model's proficiency in generating motion aligned with target meshes and maintaining shape consistency across frames. We provide a video demo for the visual inspection.

\subsection{Text-to-motion}
We compared our method with several state-of-the-art models that can generate motions given text descriptions, including T2M \cite{humanML3d}, MDM\cite{MDM}, MotionDiffuse\cite{motiondiffuse}, and Fg-T2M\cite{wang2023fg-t2m}. 

As shown in Table \ref{textmotion}, SMD outperforms other diffusion-based works in terms of FID and diversity which are indicators of motion quality, while maintaining a similar R-Precision score. 
Note that since our model takes the mesh as input, it can implicitly account for the mesh-ground interaction, even without explicit constraints to address this aspect. This is evidenced by its improved performance compared to other skeleton-based methods on physics-based metrics such as penetration and float. Examples of our generated motions are provided in the demo video.

\subsection{Action-to-motion}
We conduct this experiment for both mesh-based and skeleton-based motions, employing the settings described above. The results, summarized in Table \ref{table:actmotion}, reveal that the mesh-based method exhibits a higher accuracy upper bound, indicating its superior inherent expressive capability. This is perhaps one reason why our model outperforms the state-of-the-art diffusion-based human motion generation model. Please refer to our demo video for examples of our generated motions.

\section{Conclusion}
\label{sec:6}
We proposed SMD, a diffusion-based human motion generation model that can generate motion in the format of triangle meshes conditioned on the given text prompts and a given target mesh. We chose the spectral domain to fully exploit the meshes with fewer resource costs. Based on a Spectral-Temporal Autoencoder, our model shows great expressive ability and stability when performing the shape-conditioned generation. We expect that our method can streamline the character animation pipeline and provide new possibilities to use synthetic data to expand the human motion dataset. 
\bibliographystyle{abbrv}
\bibliography{refs}
\end{document}